\documentclass[times, review, 10pt]{elsarticle}

\usepackage{amssymb}
\usepackage{amsmath}
\usepackage{times}

\usepackage{booktabs}
\usepackage{float}
\usepackage{hyperref}
\usepackage{xcolor}
\usepackage{tabularx}
\journal{Pattern Recognition}

\begin{document}

\begin{frontmatter}



\title{A Large-Scale Dataset and a New Method for Remote Sensing Traffic Object Segmentation}



\author[label1]{Zhigang Yang} 
\ead{zgyang@mail.nwpu.edu.cn}

\author[label2]{Huiguang Yao} 
\ead{yhg2655@mail.nwpu.edu.cn}

\author[label3]{Linmao Tian} 
\ead{tianlm0808@mail.nwpu.edu.cn}

\author[label1]{Qiang Li} 
\ead{liqmges@gmail.com}

\author[label1]{Qi Wang\corref{corresponding}} 
\ead{crabwq@gmail.com}

\cortext[corresponding]{Corresponding author.}
\affiliation[label1]{organization={School of Artificial Intelligence, OPtics and ElectroNics (iOPEN), Northwestern Polytechnical University},
            addressline={Xi’an}, 
            postcode={710072}, 
            state={Shaanxi},
            country={China}}

\affiliation[label2]{organization={School of Computer Science, Northwestern Polytechnical University},
            addressline={Xi’an}, 
            postcode={710072}, 
            state={Shaanxi},
            country={China}}

\affiliation[label3]{organization={School of Software, Northwestern Polytechnical University},
            addressline={Xi’an}, 
            postcode={710072}, 
            state={Shaanxi},
            country={China}}
\begin{abstract}
Remote sensing imagery plays a crucial role in evaluating regional transportation capacity. However, existing segmentation datasets often lack diversity in object categories and scenes, limiting the ability of models to comprehensively evaluate transportation capacity in real-world scenes. To alleviate this gap, we construct a large-scale and diverse dataset for transportation object segmentation, named as NWPU-Traffic. This dataset encompass four traffic object categories (car, airplane, ship, and train) and a wide range of scenes from 49 cities across 7 countries, with instance-level annotations to ensure precise segmentation of individual objects, which bridges critical shortcomings in resolution and scene diversity in existing datasets. Leveraging this dataset, we establish a benchmark with several popular segmentation networks. Furthermore, we propose a novel segmentation method that leverages spatial-channel preserving feature interaction and an adaptive feature decoder, enabling robust segmentation across varying scales and complex environments. Extensive experiments and ablation studies validate the effectiveness of our approach. The dataset and code are publicly available at \textcolor[RGB]{237,2,140}{https://github.com/CVer-Yang/NWPU-Traffic}.

\end{abstract}



\begin{keyword}

Remote Sensing \sep Benchmark \sep Traffic Object \sep Semantic Segmentation.


\end{keyword}

\end{frontmatter}




\section{Introduction}
Extracting transportation targets within remote sensing image ~\cite{yang2025dia} regions can offer a foundation for evaluating regional economic conditions. Semantic segmentation  ~\cite{chen2024improving} of remote sensing imagery (RSI) plays a vital role in obtaining regional transportation capacity by precisely localizing and analyzing transportation objects. This ability is highly significant in applications such as smart transportation ~\cite{yang2022road}, urban planning \cite{zhang2022trans4trans}, and emergency response.

\begin{figure}[h]
\centering
\includegraphics[width=3.4in]{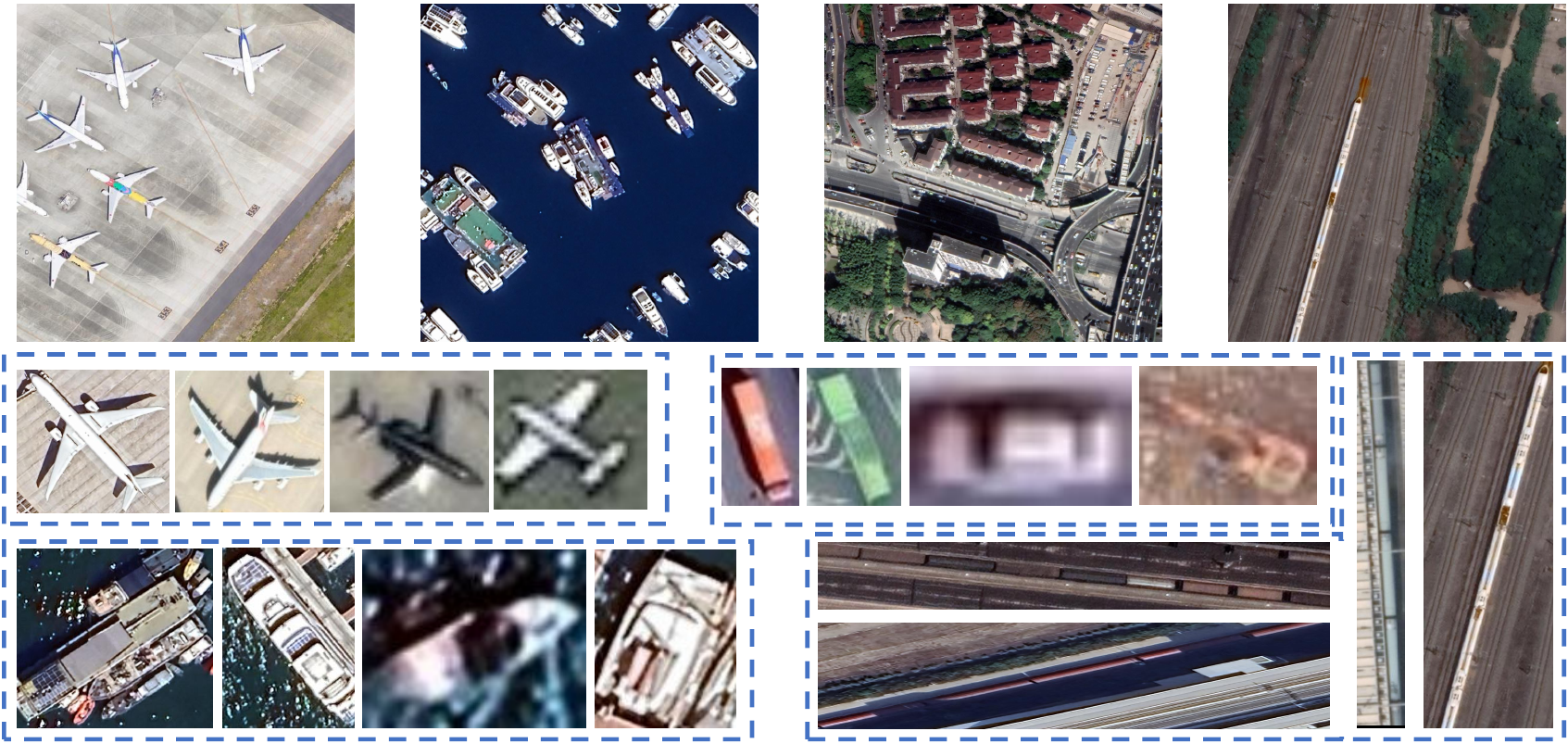}
\caption{Transportation objects observed from satellite perspectives, such as airplanes, ships, cars, and trains, exhibit significant challenges, including large-scale variations, arbitrary orientations, and diverse shapes.}
\label{fig_1}
\end{figure}

As shown in Fig. \ref{fig_1}, traffic objects in RSI exhibit unique attributes compared to other targets, including significant scale variations, arbitrary orientations, and diverse shapes. Over the past years, researchers have achieved remarkable progress in traffic object detection methods \cite{ren2024pointobb}. However, compared to detection methods, traffic object segmentation has lagged behind, primarily due to the limitations of existing datasets.

Currently, widely used traffic datasets, such as HRSC2016 \cite{liu2016ship} and UCAS-AOD \cite{zhu2015orientation} have significantly contributed to advancing intelligent traffic object recognition. However, these datasets increasingly exhibit growing limitations that hinder further progress: (1) \textbf{\textit{Limited Object Categories.}} 
Current datasets fail to provide comprehensive annotations for a diverse range of traffic object types. For example, trains—an essential mode of modern transportation—are frequently overlooked in existing RSI segmentation datasets. (2) \textbf{\textit{Limited Scene Diversity.}} Existing datasets are predominantly collected from environments like airports and harbors, which constrains model generalizability and hampers their application in complex real-world scenes. (3) \textbf{\textit{Outdated Imagery.}} The imagery in most datasets are collected before 2019, and their applicability is gradually decreasing as environmental and background conditions evolve. These limitations hinder progress in intelligent traffic applications. Therefore, constructing a comprehensive traffic object segmentation dataset that covers diverse scenes, multiple categories, and up-to-date imagery is imperative for advancing research and improving applicability in real-world scenes.

Beyond dataset limitations, existing RSI segmentation models \cite{lin2023scale} also struggle with critical architectural challenges \cite{zhang2024mrfs}: (1) \textbf{\textit{Feature Fusion Limitations.}}  Effective segmentation requires combining shallow features with positional information and deep features with semantic details. However, existing encoder-decoder architectures often lose important information during this process, which manifests as blurred boundaries and weakly activated key regions. (2) \textbf{\textit{Scale Variations.}} RSI exhibits significant variations in object scales, ranging from small vehicles to large airplanes. Most existing models rely on unified processing approaches that struggle to accommodate this scale diversity. It leads to missed detections of small objects and loss of detail in larger items. These challenges are further exacerbated in complex multi-scale scenes, where they produce incomplete and suboptimal segmentation outcomes.

To address the above challenges, we advance transportation object recognition \cite{zhan2024yolopx} in RSI through improvements in both dataset enhancements and method improvements. Specifically, we construct a comprehensive optical remote sensing image segmentation dataset, NPWU-Traffic, which consists of 1,479 images categorized into four types of transportation objects. Compared to existing datasets, NPWU-Traffic offers several significant advantages:  (1) \textbf{\textit{Comprehensive Coverage.}} This dataset includes cars, airplanes, ships, and trains, facilitating a holistic evaluation of transportation infrastructure across diverse regions. (2) \textbf{\textit{
Scene Diversity.}} The dataset spans various environments, including urban areas, rural regions, deserts, oceans, and grasslands from multiple countries,  which enhances its generalizability and robustness. (3) \textbf{\textit{Updated Imagery.}} With the majority of images captured after 2020 and Chinese data extending to 2024, which ensures alignment with modern environmental and operational conditions. Based on this dataset, we propose a novel segmentation framework that effectively integrates multi-scale features across both channel and spatial dimensions, enhancing the recognition of traffic objects at varying scales. The main contributions of this work are summarized as follows: 

\text{(1)} We present a large-scale remote sensing dataset for traffic object segmentation. This dataset encompasses diverse environments and provides a comprehensive representation of global traffic conditions. To further support advancements in RSI-based traffic object recognition, we also offer corresponding benchmarks for evaluation.

\text{(2)} We propose a spatial-channel preserving feature interaction network specifically designed for traffic object segmentation. This algorithm effectively leverages the distinctive features of multi-scale features from different stages, thereby significantly enhancing segmentation accuracy.

\text{(3)} We design a local-global feature fusion decoder that selectively combines local features with global dependencies, which can precisely segmentation of traffic objects in diverse and complex scenes.

\section{Related Works}

In this section, we review existing RSI datasets that contain traffic objects and RSI segmentation methods.
\subsection{Existing RSI Datasets Contain Traffic Object}
Current research in transportation-oriented remote sensing heavily depends on several benchmark datasets that focus specifically on cars or airplanes, such as HRSC2016, UCAS-AOD, NWPU-VHR-10 \cite{cheng2016learning}, HRRSD \cite{zhang2019hierarchical}, RSOD \cite{xiao2015elliptic}, SSDD \cite{zhang2021sar}, and LEVIR \cite{zou2017random}. Specifically, HRSC2016 serves as a dedicated maritime transportation dataset, it provides comprehensive annotations for ships and vessels across diverse sea conditions. UCAS-AOD specializes in aerial transportation targets, focusing exclusively on aircraft detection in airport environments with rotation-aware bounding boxes. The NWPU-VHR-10 dataset broadens the scope to multiple transportation categories, including ships, aircraft, and various ground vehicles, though its coverage of complex traffic scenes remains limited. For semantic segmentation of transportation infrastructure, ISAID offers large-scale annotations with three transportation-related object categories. Despite their valuable contributions to the field, existing datasets exhibit three critical limitations that impede their applicability to modern transportation research. The datasets predominantly feature isolated transportation targets (e.g., single vessels in open waters or stationary aircraft) rather than capturing realistic traffic scenes with complex object interactions and congestion patterns. Furthermore, their temporal coverage remains largely outdated, with most imagery acquired before 2018, which fails to represent contemporary transportation infrastructure and vehicle designs. Additionally, these datasets suffer from severe geographical bias, being primarily collected from limited regions without considering diverse urban layouts and regional transportation characteristics. These shortcomings significantly constrain the development of robust models for real-world applications, particularly in emerging domains such as intelligent traffic management systems and smart city infrastructure planning, where understanding complex traffic dynamics is crucial.

To address these limitations, we construct the NPWU-Traffic dataset, specifically designed for semantic segmentation tasks in remote sensing imagery. Distinguished from existing benchmarks, our dataset provides three key advantages:  (1) comprehensive coverage of transportation categories, (2) diverse scenes with complex traffic conditions, and (3) up-to-date imagery to ensure applicability to real-world applications. By offering high-quality, instance-level annotations and enhanced scene diversity, the NPWU-Traffic dataset establishes a robust benchmark for tackling contemporary research challenges and advancing transportation object segmentation in remote sensing.

\subsection{Existing RSI Segmentation Methods}

With the development of deep-learning \cite{li2023multiscale}, recent semantic segmentation advances in remote sensing imagery \cite{li2023rgb} have achieved remarkable progress in numerous applications \cite{zheng2020foreground}, including change detection \cite{zhu2024review}, infrared small-target segmentation \cite{li2024edge}, building extraction \cite{yang2024scene}, and road extraction \cite{yang2024c}. However, transportation target segmentation still faces unique challenges. A primary difficulty lies in the vast scale variations among transportation objects, from small vehicles to large aircraft and ships, which demand robust multi-scale feature representation. This is further complicated by the intricate backgrounds and frequent occlusions typical of real-world transportation scenes.
Most approaches \cite{guo2024cpp} primarily rely on multi-scale architectures \cite{huang2022scaleformer} to address these issues. For instance, Wu et al. \cite{wu2023cmtfnet} propose a network that integrates convolution and Transformers. By using a multi-scale attention fusion module to adaptively combine deep and shallow features, it significantly enhances semantic segmentation of high-resolution remote-sensing images. To better model long-range dependencies, Zeng et al. \cite{zeng2024multiscale} develop multi-scale interaction modules and a Transformer-based decoder. They also introduce a scale-aware fusion module to improve supervision ability. Similarly, Li et al. \cite{10969832} combine convolution, self-attention mechanisms, and Mamba models. They introduce novel global-local visual fusion blocks to embed local information into global information through feature similarity.
Typically, these methods adjust resolution through pooling and upsampling or manipulate the channel dimension. Although they achieve acceptable performance in some cases, they face inherent limitations in preserving crucial spatial information during feature fusion. Moreover, existing techniques often introduce fusion artifacts and noise when integrating features from different levels, which will lead to reduced segmentation quality at object boundaries and fine structures. To overcome these issues, we introduce a novel multi-scale feature fusion framework that preserves spatial and channel information during skip connections. Additionally, we incorporate a lightweight global-local feature fusion module in the decoder stage to combine local features with global semantic information.

\section{The NWPU-Traffic Dataset Construction}\label{sec3}
In this section, we provide a comprehensive overview of the dataset and conduct a thorough dataset analysis.

\subsection{Dataset Collection}
\begin{figure*}[ht]
\centering
\includegraphics[width=4.83in]{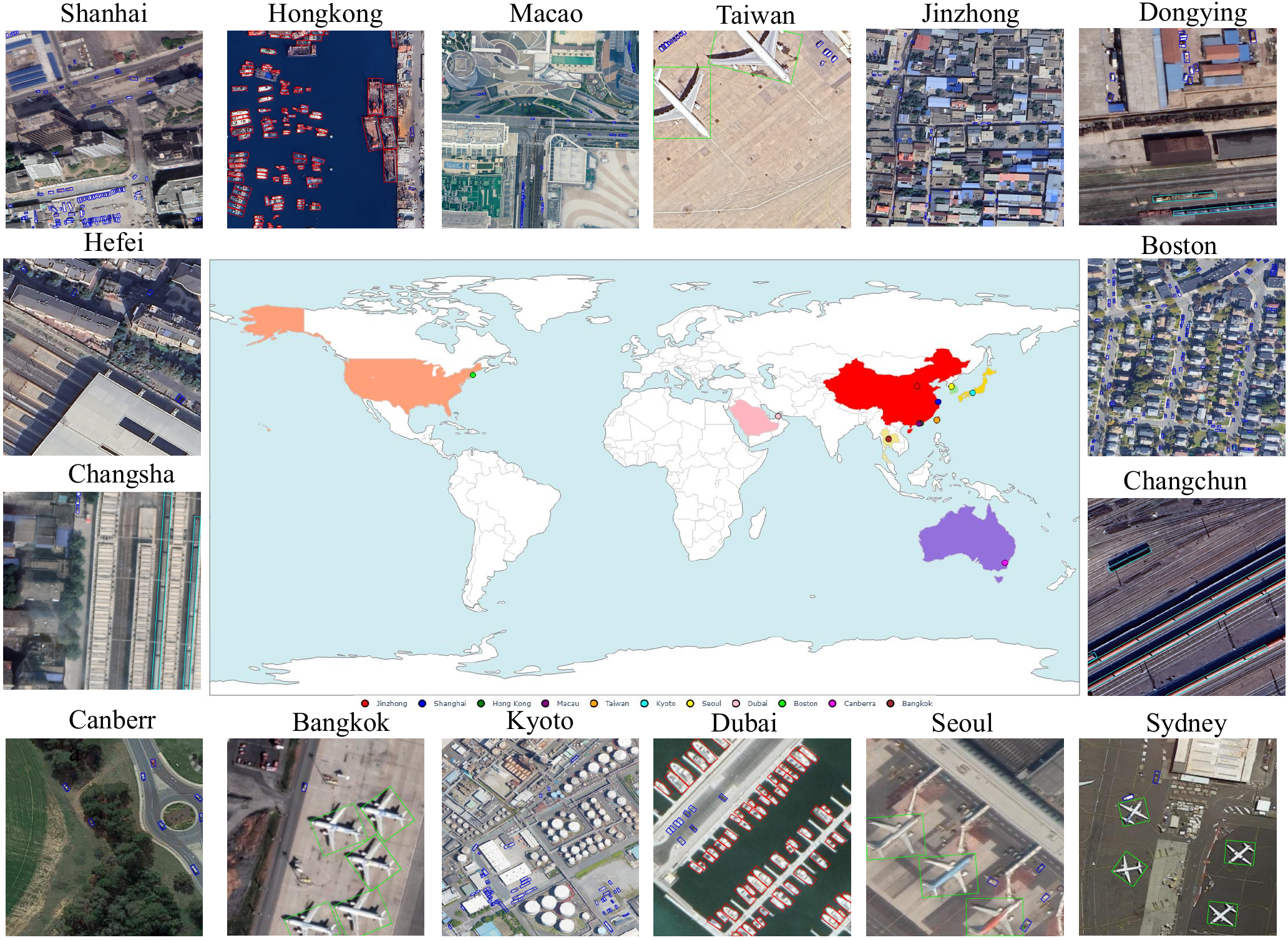}
\caption{We select typical traffic scenes from seven countries globally for traffic target labeling, including China, USA, Japan, South Korea, Thailand, the United Arab Emirates, and Australia. Furthermore, we transform segmentation masks into rotated bounding boxes to enhance visualization.}
\label{fig_2}
\end{figure*}

As shown in Fig. \ref{fig_2}, the dataset constructed in this study encompasses representative geographical diversity and a variety of scenes from around the world. We systematically collected image data from 49 cities across 7 countries, covering various ecosystems such as urban landscapes, rural environments, marine areas, and desert zones. To ensure the broad applicability of the dataset in real-world applications, we focus on collecting data from typical traffic scenes, including key transportation facilities such as urban centers, airports, ports, and railway stations.

\subsection{Dataset Annotation}
To ensure accuracy and consistency in data annotation, a team of three experienced annotators with expertise in remote sensing image interpretation is assembled and trained on the dataset standards. The process includes two stages: detailed annotation and quality inspection. During the annotation stage, the annotators label the image data with precise location and category information for transportation objects. Following this, a team of experts conducts a comprehensive review of the annotated samples to evaluate object boundary accuracy, correctness of category annotations, and to identify any omissions or misannotations. In addition, we transformed masks into rotated bounding boxes via code, ultimately offering both traffic target masks and rotated bounding box outputs.

\subsection{Dataset Analysis}

\begin{figure*}[ht]
\centering
\includegraphics[width=4.83in]{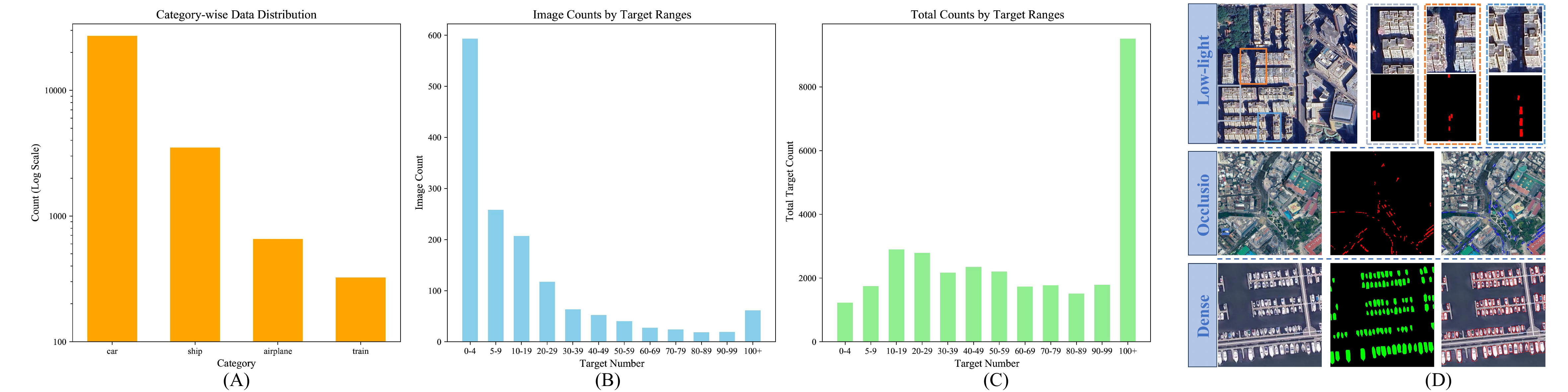}
\caption{(A), (B), and (C) demonstrate the image count by category, target number distribution, and cumulative target counts per range, respectively.}
\label{fig_3}
\end{figure*}

In this subsection, we analyze the characteristics of the NWPU-Traffic dataset and compare it with related remote sensing segmentation datasets.

\textbf{Instance Count:} As illustrated in Fig. \ref{fig_3}(A), the dataset consists of 1,479 images with ground sample distances ranging from 0.12m to 0.5m per pixel. 
It contains a total of 31,628 object instances, with an average of 21.38 objects per image. The target distribution reveals that cars represent the largest proportion, while trains are comparatively fewer. In comparison to existing datasets, NPWU-Traffic offers a more comprehensive and densely populated variety of transportation objects. This diversity facilitates more robust model training and enables effective performance evaluation in multi-object scenes.

\textbf{Instance Distribution Analysis:} 
We conduct a statistical analysis of the dataset from two perspectives: image instance distribution and target quantity distribution. As shown in Fig. \ref{fig_3}(B), images with 0–4 instances constitute the largest proportion, approximately 39.61\%. Furthermore, the dataset exhibits a high-density distribution characteristic, with many images containing a significant number of targets. In Fig. \ref{fig_3}(C), the total number of targets within different quantity ranges is analyzed. The results reveal that a substantial portion of the targets originates from high-density images, accounting for approximately 30.05\% of the total instances in the dataset. This distribution highlights that the NPWU-Traffic dataset includes both low-density and high-density scenes. It offers a realistic representation of transportation object distributions in remote sensing imagery.  This balance ensures that the dataset can effectively support segmentation tasks under varying target densities.

\subsection{Typical Scene Analysis}
\begin{figure}[ht]
\centering
\includegraphics[width=3.0in]{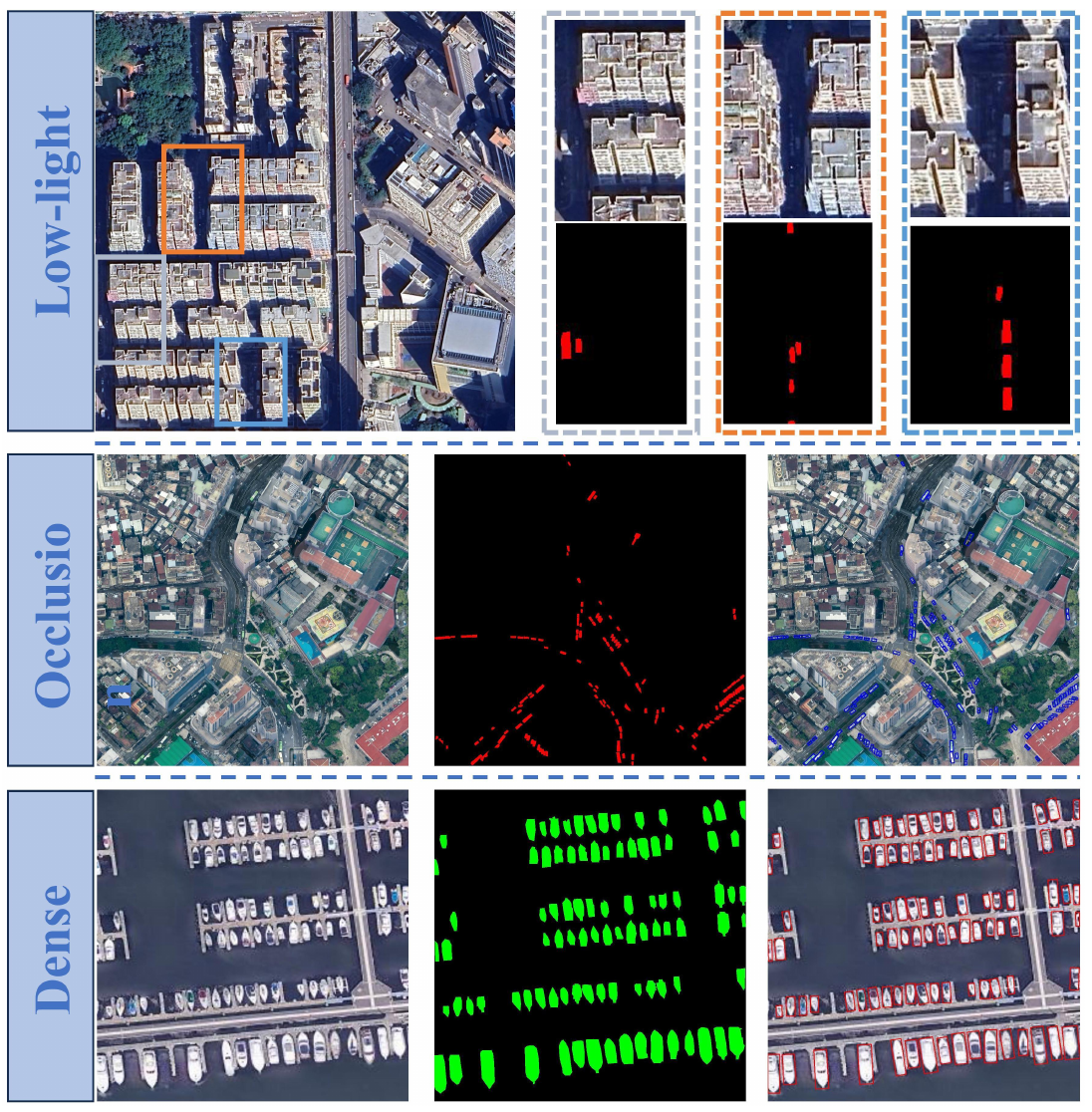}
\caption{We show some representative scenes of the NWPU-Traffic and convert the segmentation masks of traffic targets into rotated boxes for visualization.}
\label{fig_4}
\end{figure}

During the data collection process, three representative types of scenes are prioritized, with sample images as shown in the Fig. \ref{fig_4}:

\textbf{Low-Light Scenes:} These scenes focus on scenes where vehicles are positioned in shadowed areas, such as beneath urban buildings. Low-light scenes aim to evaluate model performance under challenging lighting conditions, which ensures adaptability to real-world environments.

\textbf{Occlusion Scenes:} These scenes involve transportation objects partially obscured by elements such as high-rise buildings, bridges, or trees. The inclusion of occlusion scenes enhances model robustness in handling complex environments with partial visibility.

\textbf{Dense Scenes:} These scenes are characterized by a high density of targets, commonly observed in areas such as parking lots and docks. Dense scenes serve to assess the ability of model to accurately recognize and segment objects in crowded and cluttered settings.

\subsection{Comparison with Existing Dataset}
\begin{table*}[htbp]
\centering
\caption{Comparison between remote sensing datasets containing traffic objects. NWPU-Traffic serves as a valuable supplement to existing remote sensing datasets, offering polygon segmentation masks for four traffic object categories.}
\label{tab1}
\resizebox{\textwidth}{!}{
\begin{tabular}{lcccccccc}
\toprule
\textbf{Dataset} & 
\textbf{Bounding box} & 
\textbf{Segmentation mask} & 
\textbf{Global} & 
\textbf{Traffic categories} & 
\textbf{Images} & 
\textbf{Instances} & 
\textbf{Year} \\
\midrule
TAS \cite{heitz2008learning} & horizontal & \texttimes & \texttimes & 2 & 30 & 1,319 & 2008 \\
UCAS-AOD \cite{zhu2015orientation} & horizontal & \texttimes & \texttimes & 2 & 910 & 6,029 & 2015 \\
NWPU VHR-10 \cite{cheng2016learning} & horizontal & \texttimes & \texttimes & 3 & 800 & 1,536 & 2016 \\
COWC \cite{mundhenk2016large} & center-point & \texttimes & \texttimes & 1 & 53 & 32,716 & 2016 \\
VEDAI \cite{razakarivony2016vehicle} & oriented & \texttimes & \texttimes & 3 & 1,200 & 3,700 & 2016 \\
HRSC2016 \cite{liu2016ship} & oriented & \texttimes & \texttimes & 1 & 1,061 & 2,976 & 2016 \\
LEVIR \cite{zou2017random} & horizontal & \texttimes & \texttimes & 2 & 21,952 & 7,749 & 2017 \\
xView \cite{lam2018xview} & horizontal & \texttimes & \texttimes & 1 & 1,127 & 1,000,000 & 2018 \\
DOTA \cite{xia2018dota} & oriented & \texttimes & \texttimes & 3 & 1,869 & 100,450 & 2018 \\
HRRSD \cite{zhang2019hierarchical} & horizontal & \texttimes & \texttimes & 3 & 26,722 & 13,541 & 2019 \\
Airbus Ship \cite{AirbusShipDetectionChallengeDataset} & polygon & \checkmark & \texttimes & 1 & 131,000 & 192,000 & 2019 \\
iSAID \cite{waqas2019isaid} & polygon & \checkmark & \texttimes & 3 & 1,869 & 118,188 & 2019 \\
SSDD \cite{zhang2021sar} & horizontal & \checkmark & \texttimes & 1 & 1,160 & 2,456 & 2021 \\
NWPU-Traffic (Ours) & polygon & \checkmark & \checkmark & 4 & 1,479 & 31,628 & 2025 \\
\bottomrule
\end{tabular}
}
\end{table*}

As shown in Table \ref{tab1}, current remote sensing traffic target datasets mainly use detection-level bounding box annotations, with only a few including detailed instance-level annotations. In contrast, the NWPU-Traffic dataset is the first remote sensing dataset to comprehensively cover four categories of traffic targets, thus filling a significant gap in the field. Existing datasets usually focus on single geographic regions, such as specific cities or countries, which limits the diversity and generalization of the traffic targets. However, the NWPU-Traffic dataset includes global multi-regional scenes, greatly improving its representativeness and applicability. In terms of timeliness, most data are collected before 2020, and few new datasets have been introduced in recent years. Consequently, these datasets fail to reflect recent developments in transportation infrastructure, particularly the adoption of new energy vehicles such as electric buses. In contrast, our dataset is derived from up-to-date satellite imagery spanning 2020 to 2024, better providing a more accurate representation of current urban environments. In summary, the NWPU-Traffic dataset has remarkable advantages in annotation granularity, target systematicness, geographic coverage, and timeliness. It is an important data supplement for research on remote sensing traffic target segmentation and detection.

\section{Proposed Method} \label{sec4}

\subsection{Overview}

\begin{figure*}[htb]
\centering
\includegraphics[width=4.83in]{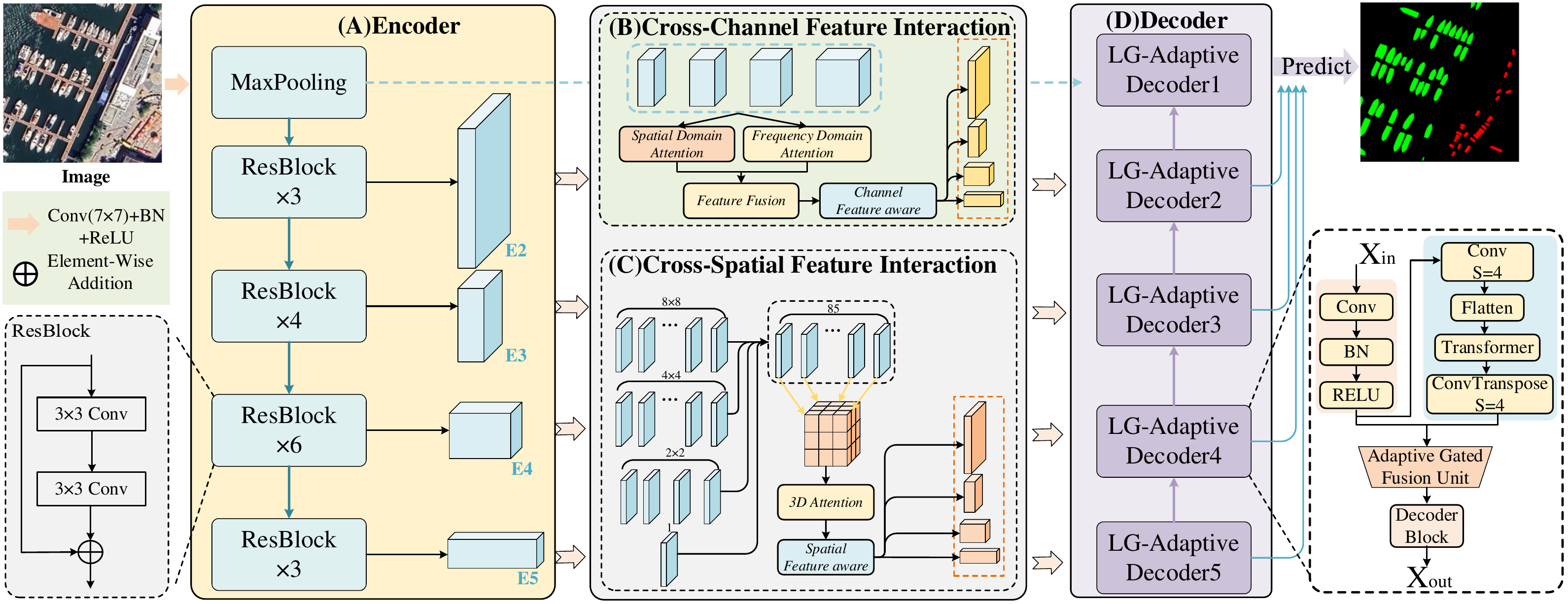}
\caption{The overall framework of the proposed CSPNet, and it is divided into four sections: the encoder, spatial-preserved feature interaction module, channel-preserved feature interaction module, and local-global feature fusion Decoder.}
\label{fig_5}
\end{figure*}

The overall framework of the proposed model is illustrated in Fig. \ref{fig_5}. The model utilizes a ResNet-34 encoder to extract five hierarchical feature maps, denoted as E1, E2, E3, E4, and E5. These feature maps are further refined through two specialized modules: the Spatial-Preserved Feature Interaction Module (SPFIM) and the Channel-Preserved Feature Interaction Module (CPFIM). The SPFIM maintains spatial positional information during cross-scale interactions, while the CPFIM ensures channel-wise consistency. It enables efficient and robust feature fusion across scales. These refined feature maps and decoder features are then fed into the Local-Global Feature Fusion Decoder (LGFFD), which adaptively integrates long-range contextual information with fine-grained local details. A detailed explanation of each component of the model is provided in the following sections.

\subsection{Channel-Preserved Feature Interaction Module} 

\begin{figure*}[ht]
\centering
\includegraphics[width=4.83in]{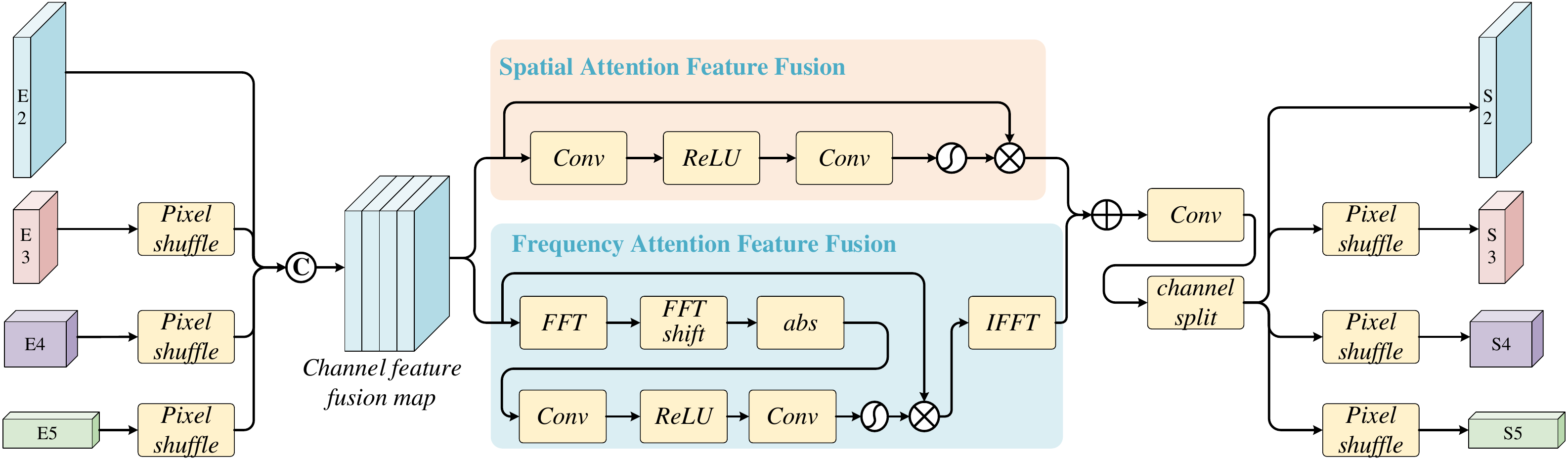}
\caption{Details of channel-preserved feature interaction module.}
\label{fig_6}
\end{figure*}

Preserving channel information is crucial for enhancing the performance of segmentation models. Channel information contains rich semantic features, and its loss may result in semantic ambiguity and reduced discrimination between different traffic objects. To address this issue, we propose a CPFIM that combines channel and frequency-domain weighting mechanisms. This module allows efficient feature fusion while maintaining channel integrity and utilizing complementary information from both spatial and frequency domains. As shown in Fig. \ref{fig_6}, the proposed CPFIM consists of four parts.

\textbf{\textit{Resolution Unification.}} To unify feature map resolutions across different scales, we employ the pixel shuffle operation to compress channel information into spatial dimensions. Let the input feature map be denoted as $
{\mathbf{X}}_i \in \mathbb{R}^{C_i \times H_i \times W_i}$, where \(C_i\), \(H_i\), and \(W_i\) represent the channels, height, and width of the \(i\)-th feature map, respectively. The pixel shuffle operation rearranges the spatial and channel information and yields $
\overline{\mathbf{X}}_i \in \mathbb{R}^{C_i' \times H_i' \times W_i'}
$, i.e.,  
\[
(C_i, H_i, W_i) \longrightarrow \left(C_i' = \frac{C_i}{s^2}, H_i' = s \cdot H_i, W_i' = s \cdot W_i\right),
\]  
and \(s\) denotes the shuffle factor. This ensures resolution alignment for coherent feature interactions across different scales. These features $\mathbf{X}_i$ are then concatenated in the channel dimension to generate the channel fused map $X_{fuse}$.

\textbf{\textit{Channel Domain Weighting.}} In the channel domain, the feature map $X_{fuse}$ is refined through convolution and activation operations, which compress the channel dimension to generate compact representations. The channel dimension is subsequently restored, and a sigmoid activation function produces adaptive channel-wise weights ${W}_c$, i.e.,
\[
{X}_c = {X_{fuse}} \odot {W}_c,
\]
where \(\odot\) denotes element-wise multiplication, dynamically emphasizing significant channels.

\textbf{\textit{Frequency Domain Weighting.}}  To capture global correlations, feature maps are transformed into the frequency domain using the Fast Fourier Transform. Subsequently, the frequency features are normalized, and we extract spectral amplitudes as input. A convolution, activation, convolution, and sigmoid operation is applied to generate frequency-domain weights, enabling the adjustment of frequency-transformed feature maps, i.e.,
\[
G(x) = F({X_{fuse}}) \odot {K_{1 \times 1}}(R({K_{3 \times 3}}(\left| {{X_{fuse}}} \right|)))\,
\]
where $K_{3 \times 3}( \cdot )$ is convolution operation with kernal size of $3\times3$, $R(\cdot)$ is ReLU operation, and $F(\cdot)$ is fast fourier transform operation. Finally, the Inverse Fast Fourier Transform is applied to convert the frequency-weighted features into the spatial domain feature $X_{f}$.

\textbf{\textit{Feature Division.}} The refined feature maps ${X}_c$ and ${X}_f$ are fused via an element-wise addition operation to form a unified representation.  Then, a convolution operation is applied to fuse the concatenated features. Finally, a pixel unshuffle operation restores the original spatial and channel dimensions, producing the final output, i.e.,
\[
{CP}_i = \mathcal{P}(\mathcal{S}({K_{1 \times 1}}([{X}_c, {X}_f]))),
\]
where $\mathcal{P}(\cdot)$ is the pixel unshuffle operation, $\mathcal{S}(\cdot)$ is the split operation, \([\cdot, \cdot]\) denotes feature concatenation operation. These cross-channel interaction features are input into the decoder for semantic information enhancement.

\subsection{Spatial-Preserved Feature Interaction Module} 

\begin{figure*}[ht]
\centering
\includegraphics[width=4.83in]{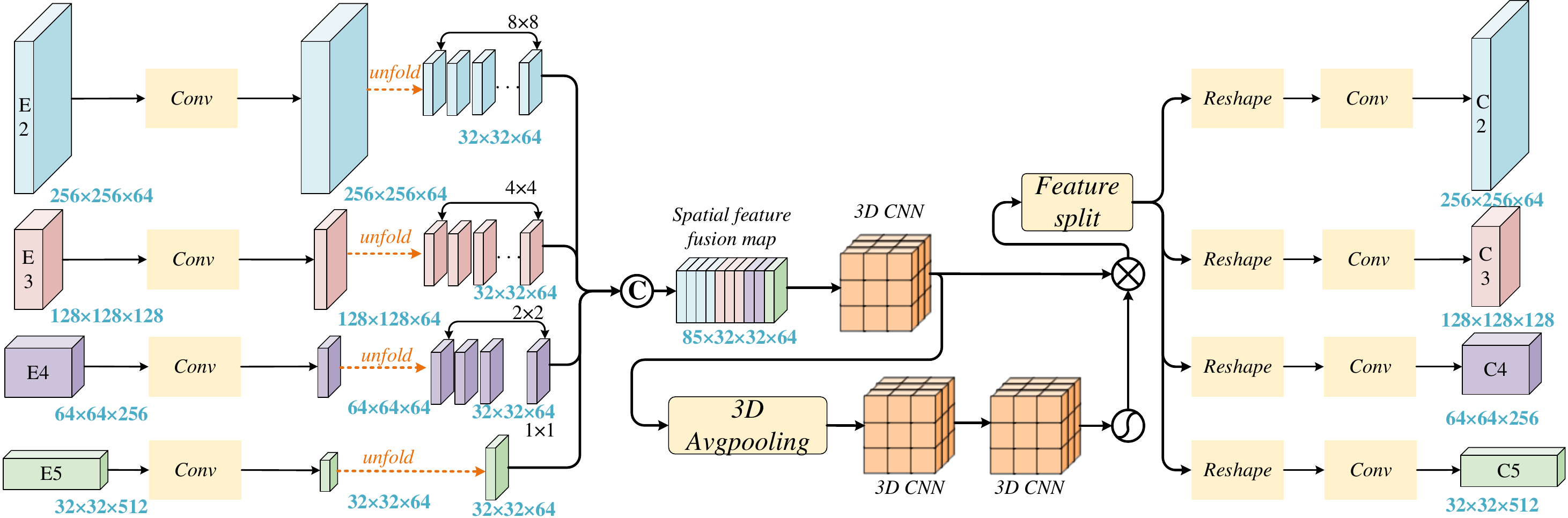}
\caption{Details of spatial-preserved feature interaction module.}
\label{fig_7}
\end{figure*}

The fusion of multi-stage feature maps is crucial for enabling cross-scale information integration in segmentation tasks. However, traditional operations such as concatenation or summation often result in the loss of positional information, which is critical for accurately locating and identifying transportation targets. To address this challenge, we propose a SPFIM, which facilitates efficient cross-scale feature fusion while ensuring maximal retention of positional information across feature maps. As illustrated in Fig. \ref{fig_7}, the SPFIM consists of three stages.

\textbf{\textit{Feature Alignment.}} The process begins by standardizing the channel dimensions of all input feature maps through a convolution operation, which aligns all feature maps to 64 channels. The channel-aligned feature maps are given by
\[
\widetilde{{X}}_i  = K_{1 \times 1}({X}_i), \quad \widetilde{{X}_i} \in \mathbb{R}^{64 \times H_i \times W_i}.
\]
Next, an unfold operation decomposes the input feature maps into smaller patches, matching the resolution of the coarsest feature map ${F}_5$. Each feature map is divided into patches of sizes \(8 \times 8\), \(4 \times 4\), \(2 \times 2\), and \(1 \times 1\), with all patches being adjusted to a uniform spatial size. These patches are then concatenated along the sequence dimension, which results in a fused sequence feature map $
{F}_\text{sequence} \in \mathbb{R}^{85 \times 32 \times 32 \times 64}$.

\textbf{\textit{Spatial Domain Weighting.}} To enable effective cross-scale feature interaction, a 3D convolution operation is applied to the fused feature map. This operation captures dependencies and relationships between features at different scales, which facilitates comprehensive and robust feature integration. To further refine the fused feature map and emphasize critical information, a sequence channel attention mechanism is introduced. This mechanism dynamically weights channels to prioritize essential features. The process begins with 3D average pooling, which compresses the spatial information of the feature map into global channel features ${F}_\text{avg}$. A 3D convolution operation reduces the channel dimension, followed by a ReLU activation for non-linear transformation.
And another 3D convolution restores the channel dimension, it generates adaptive channel-wise weights $W_c$ through a sigmoid activation. These weights are applied to the fused feature map via element-wise multiplication to emphasize critical features, i.e.,
\[
{F}_\text{s} = {F}_\text{sequence} \odot {W}_c.
\]

\textbf{\textit{Feature Division.}} The refined feature map is divided back into four feature map sets along the sequence dimension. A reshape operation restores these patches to their original spatial resolutions, and a convolution operation adjusts the channel dimensions as needed,i.e.,
\[
{SP}_i = {K_{1 \times 1}}(\mathcal{R}(\mathcal{S}({F}_\text{s}))),
\]
where $\mathcal{R}(\cdot)$ is the Reshape operation.

\subsection{Local-Global Feature Fusion  Decoder} 

In semantic segmentation, the decoding stage is pivotal for reconstructing high-resolution predictions with both semantic richness and spatial precision. While conventional decoders based solely on convolution or Transformers have demonstrated competence, their inherent architectural biases introduce fundamental trade-offs: convolution preserve local structural details but suffer from limited receptive fields, whereas Transformers excel at modeling global context while often compromising fine-grained spatial information. To address these challenges, we propose the Local-Global Feature Fusion Decoder, which combines the complementary strengths of convolution and Transformers through a dynamic gating mechanism to adaptively fuse local and global features. The proposed decoder consists of three core components.

The convolution branch focuses on extracting local contextual information and fine-grained spatial features through a 3×3 convolution, which reduces the channel dimension to C/4 for computational efficiency, followed by Batch Normalization and ReLU activation to enhance feature robustness. In parallel, the Transformer branch is designed to capture global semantic relationships and long-range dependencies. The process begins with a depthwise convolution to reduce the spatial dimensions of the input feature map, followed by a transformer block that models global dependencies across the entire feature map.

To optimally integrate these complementary representations, we introduce a gated fusion module that dynamically balances their contributions through learnable spatial weights. These weights $W$ are generated via a 1×1 convolution with sigmoid activation. The fused feature map is then calculated as
\[
{F}_{\text{fuse}} = {W} \odot {F}_{\text{local}} + (1 - {W}) \odot {F}_{\text{global}}.
\]
The fused features are then upsampled using a transposed convolution, refined through batch normalization and ReLU, and finally projected to the target segmentation space via a convolutional layer. The aggregated feature map is then fused with the semantically enhanced features ${CP}_i$ from the CPFIM  and the spatially enhanced features ${SP}_i$ from the SPFIM through element-wise summation, ultimately producing the decoder output feature map for this stage.

\section{Experiments}

\subsection{Dataset Setup and Evaluation Metrics}

The dataset is divided into train, validation, and test sets using a 7:1:2 ratio based on the sample distribution across different regions. This division ensures the representativeness and balance of the data, avoiding regional biases. Such a split strategy enables all regions to be well covered during the training process, improving the robustness and generalization of the model. The following metrics are selected: Intersection over Union (IoU) for each transportation target, Mean Intersection over Union (MIoU), Overall Accuracy (OA), and the F1-score. These metrics provide a comprehensive assessment of the model performance from multiple perspectives.

\subsection{Experimental Settings}
All experiments are conducted using PyTorch and trained on an NVIDIA GTX 4090 GPU. We employ the AdamW optimizer initialized with a base learning rate of 6e-4 and a cosine annealing learning rate scheduler to facilitate stable convergence. The combination of binary cross entropy with Dice coefficient is used as the loss function. All input images are randomly cropped to 512×512 patches and augmented with scaling [0.5,0.75,1.0,1.25,1.5], flipping, and rotation. The batch size is 4. During test, we implement comprehensive test-time augmentation, incorporating both flip-based and multi-scale strategies, where final predictions are obtained through averaging outputs across all augmented variants to maximize segmentation robustness.

\subsection{Comparison with Existing Methods}

To evaluate the performance of the proposed model, we compare it with several widely adopted segmentation methods, including DeepLabv3+ \cite{chen2018encoder}, RSI segmentation approaches such as ABCNet \cite{li2021abcnet}, UNetFormer \cite{wang2022unetformer}, CMLFormer \cite{wu2024cmlformer}, and GCBNet \cite{zhu2021global}, the recent segmentation framework ScaleFormer \cite{huang2022scaleformer}, and the FSEL \cite{sun2025frequency} method.

\begin{table*}[htb]
    \caption{Results on NWPU Traffic dataset. The best results are bold and the second-best results are underlined.}
    \centering
    \resizebox{\textwidth}{!}{
    \begin{tabular}{c|cc|cccccccc}  
    \toprule
        \textbf{Method} & 
        \textbf{FLOPs}\textsuperscript{\textcolor{red}{$\downarrow$}} (G) &  
        \textbf{Params}\textsuperscript{\textcolor{red}{$\downarrow$}} (M) &  
        \textbf{Background}\textsuperscript{\textcolor{blue}{$\uparrow$}} & 
        \textbf{Airplane}\textsuperscript{\textcolor{blue}{$\uparrow$}} & 
        \textbf{Car}\textsuperscript{\textcolor{blue}{$\uparrow$}} & 
        \textbf{Ship}\textsuperscript{\textcolor{blue}{$\uparrow$}} & 
        \textbf{Train}\textsuperscript{\textcolor{blue}{$\uparrow$}} & 
        \textbf{mIoU}\textsuperscript{\textcolor{blue}{$\uparrow$}} & 
        \textbf{OA}\textsuperscript{\textcolor{blue}{$\uparrow$}} & 
        \textbf{F1-score}\textsuperscript{\textcolor{blue}{$\uparrow$}} \\
    \midrule
        Deeplabv3+ & 185.6 & 41.2 & \underline{98.66} & \textbf{86.02} & 45.06 & 74.66 & \textbf{56.59} & 72.19 & \underline{98.69} & 82.34 \\
        ABCNet & 98.3 & 28.5 & 97.86 & 77.63 & 29.57 & 64.75 & 17.28 & 57.42 & 97.89 & 68.01 \\
        UNetFormer & 112.7 & 30.1 & 97.98 & 78.91 & 32.20 & 66.73 & 22.28 & 59.62 & 98.01 & 70.48 \\
        GCBNet & 156.8 & 35.7 & 98.48 & 85.03 & 48.28 & 75.18 & 40.55 & 69.50 & 98.51 & 79.95 \\
        Scaleformer & 134.2 & 32.8 & 98.22 & 74.45 & 33.49 & 64.74 & 46.37 & 63.45 & 98.25 & 75.32 \\
        CMLFormer & 148.5 & 38.4 & 98.64 & \underline{85.67} & \underline{48.82} & \underline{78.43} & 53.22 & \underline{72.95} & 98.68 & \underline{83.18} \\
        FSEL & 126.9 & 31.5 & 98.47 & 84.54 & 30.88 & 72.35 & 50.14 & 67.28 & 98.50 & 80.51 \\
        CSPNet & \textbf{118.7} & \textbf{29.6} & \textbf{98.73} & 84.52 & \textbf{48.91} & \textbf{80.07} & \underline{55.30} & \textbf{73.50} & \textbf{98.76} & \textbf{83.36} \\
    \bottomrule
    \end{tabular}
    }
    \label{tab2}
\end{table*}

\begin{table*}[htb]
    \caption{Results on NWPU Traffic dataset. The best results are bold and the second-best results are underlined.}
    \centering
    \resizebox{\textwidth}{!}{
    \begin{tabular}{c|cccccccc}
    \toprule
        \textbf{Method} & 
        \textbf{Background}\textsuperscript{\textcolor{blue}{$\uparrow$}} & 
        \textbf{Airplane}\textsuperscript{\textcolor{blue}{$\uparrow$}} & 
        \textbf{Car}\textsuperscript{\textcolor{blue}{$\uparrow$}} & 
        \textbf{Ship}\textsuperscript{\textcolor{blue}{$\uparrow$}} & 
        \textbf{Train}\textsuperscript{\textcolor{blue}{$\uparrow$}} & 
        \textbf{mIoU}\textsuperscript{\textcolor{blue}{$\uparrow$}} & 
        \textbf{OA}\textsuperscript{\textcolor{blue}{$\uparrow$}} & 
        \textbf{F1-score}\textsuperscript{\textcolor{blue}{$\uparrow$}} \\
    \midrule
        Deeplabv3+ & \underline{98.66} & \textbf{86.02} & 45.06 & 74.66 & \textbf{56.59} & 72.19 & \underline{98.69} & 82.34 \\
        ABCNet & 97.86 & 77.63 & 29.57 & 64.75 & 17.28 & 57.42 & 97.89 & 68.01 \\
        UNetFormer & 97.98 & 78.91 & 32.20 & 66.73 & 22.28 & 59.62 & 98.01 & 70.48 \\
        GCBNet & 98.48 & 85.03 & 48.28 & 75.18 & 40.55 & 69.50 & 98.51 & 79.95 \\
        Scaleformer & 98.22 & 74.45 & 33.49 & 64.74 & 46.37 & 63.45 & 98.25 & 75.32 \\
        CMLFormer & 98.64 & \underline{85.67} & \underline{48.82} & \underline{78.43} & 53.22 & \underline{72.95} & 98.68 & \underline{83.18} \\
        FSEL & 98.47 & 84.54 & 30.88 & 72.35 & 50.14 & 67.28 & 98.50 & 80.51 \\
        CSPNet & \textbf{98.73} & 84.52 & \textbf{48.91} & \textbf{80.07} & \underline{55.30} & \textbf{73.50} & \textbf{98.76} & \textbf{83.36} \\
    \bottomrule
    \end{tabular}
    }
    \label{tab2}
\end{table*}

The results in Table \ref{tab2} demonstrate that CSPNet achieves state-of-the-art performance across three key metrics: OA, mIoU, and F1-score, which shows the effectiveness of the proposed modules. The model demonstrates particular strength in background segmentation, which indicates excellent generalization across diverse scenarios. For challenging small object categories, CSPNet attains superior IoU scores of 48.91\% and 80.07\% for cars and ships respectively, surpassing the closest competitors by margins of 0.09\% and 1.64\%. Comparative analysis shows DeepLabv3+ delivers strong results for airplanes and trains but struggles with cars. This limitation may stem from architectural constraints. The spatial pyramid dilated convolution successfully captures multi-scale context, but the global average pooling operation causes adjacent object features to merge, particularly detrimental for dense small targets such as cars and ships. CMLFormer achieves competitive results through combined local and global feature extraction, yet the absence of dedicated multi-scale fusion mechanisms creates a measurable performance gap. Scaleformer attempts multi-scale integration but fails to properly align features across scales, which limits its effectiveness. Overall, CSPNet delivers competitive performance, especially on challenging categories such as cars and ships. These comprehensive results highlight unique advantages of CSPNet in multi-scale feature interaction and fusion. The architecture delivers consistent performance gains across all object categories while specifically addressing the segmentation challenges posed by small, dense targets in complex traffic scenes. The superior performance further highlights the advantages of the proposed modules in addressing the limitations of existing approaches and achieving precise segmentation results across a wide range of traffic object and environmental complexities.


\subsection{Visual Results}
\begin{figure*}[h]
\centering
\includegraphics[width=4.83in]{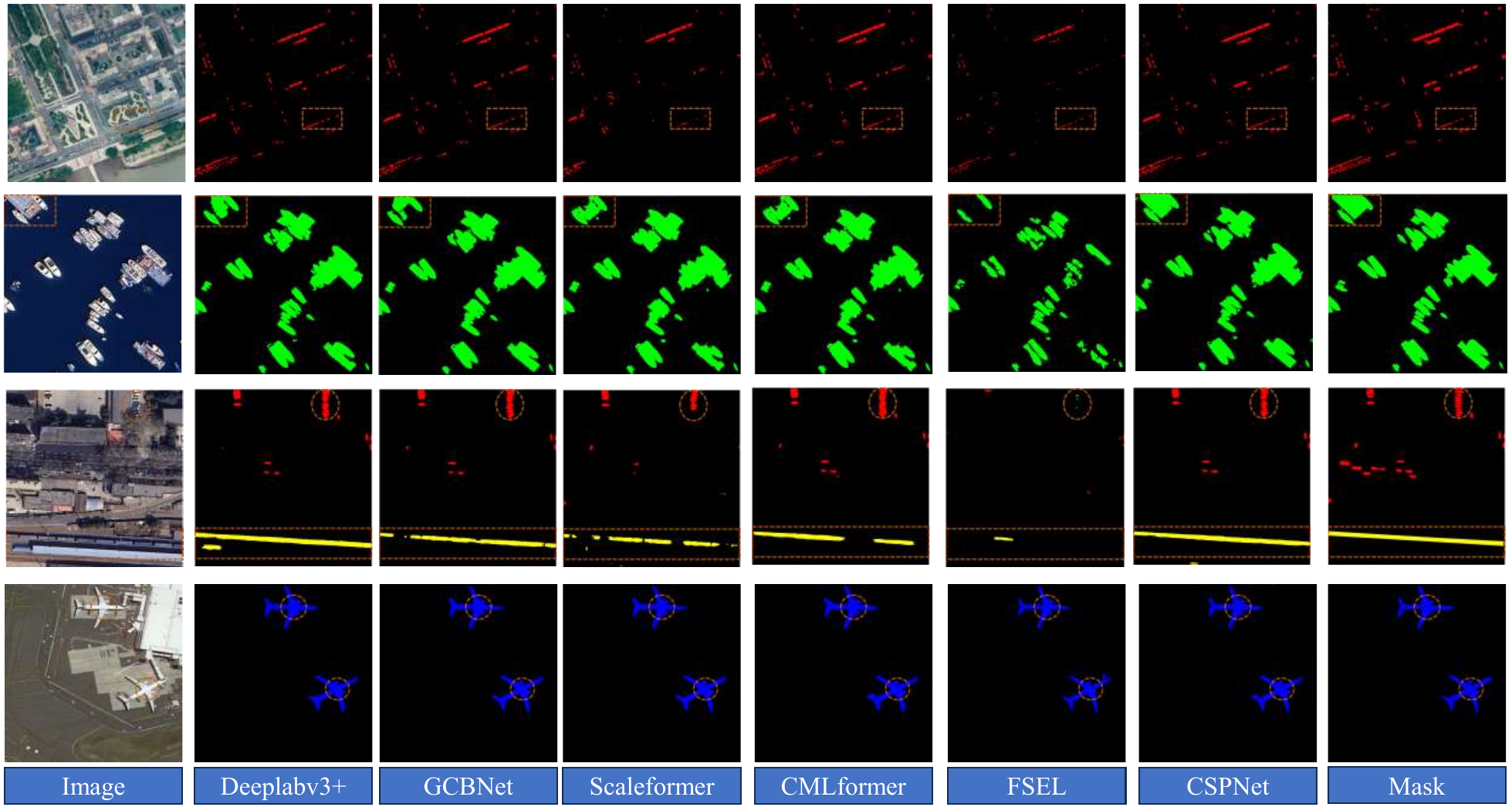}
\caption{Visualization results of the model on the NWPU-Traffic dataset.}
\label{fig_9}
\end{figure*}

Fig. \ref{fig_9} presents a qualitative comparison of segmentation performance among different models on the NWPU-Traffic dataset. Based on descending mIoU scores, we select the top six performing models for comparative visualization and analysis. The results show that CSPNet attains superior performance in object localization and boundary segmentation accuracy. Consistently outperforming alternative approaches, the proposed model shows clear advantages in processing transportation objects with diverse scales and complex structures. Specifically, CSPNet displays remarkable ability in segmenting large-scale transportation targets. For maritime scenes shown in the second row, CSPNet produces significantly more complete and accurate ship segmentation results compared to other methods. Similarly, in railway scenes illustrated in the third row, the model maintains consistent boundary delineation for elongated objects such as trains, demonstrating robustness when handling complex geometric shapes. Moreover, CSPNet excels at preserving fine-grained details. The fourth row emphasizes the model capacity to capture intricate airplane features with precise edge definition, in contrast to the fragmented or blurred boundaries generated by competing approaches. This highlights the CSPNet advantage in retaining delicate structural information. The comparative visualization confirms that the hierarchical feature fusion mechanism in CSPNet effectively balances contextual understanding with spatial accuracy. This dual ability enables the model to generate segmentation masks that preserve both global structural coherence and local detail precision, it sets a new standard for transportation scene analysis on the NWPU-Traffic dataset.

\subsection{Complex Scenes Visual Results
}

\begin{figure}[h]
\centering
\includegraphics[width=3.38in]{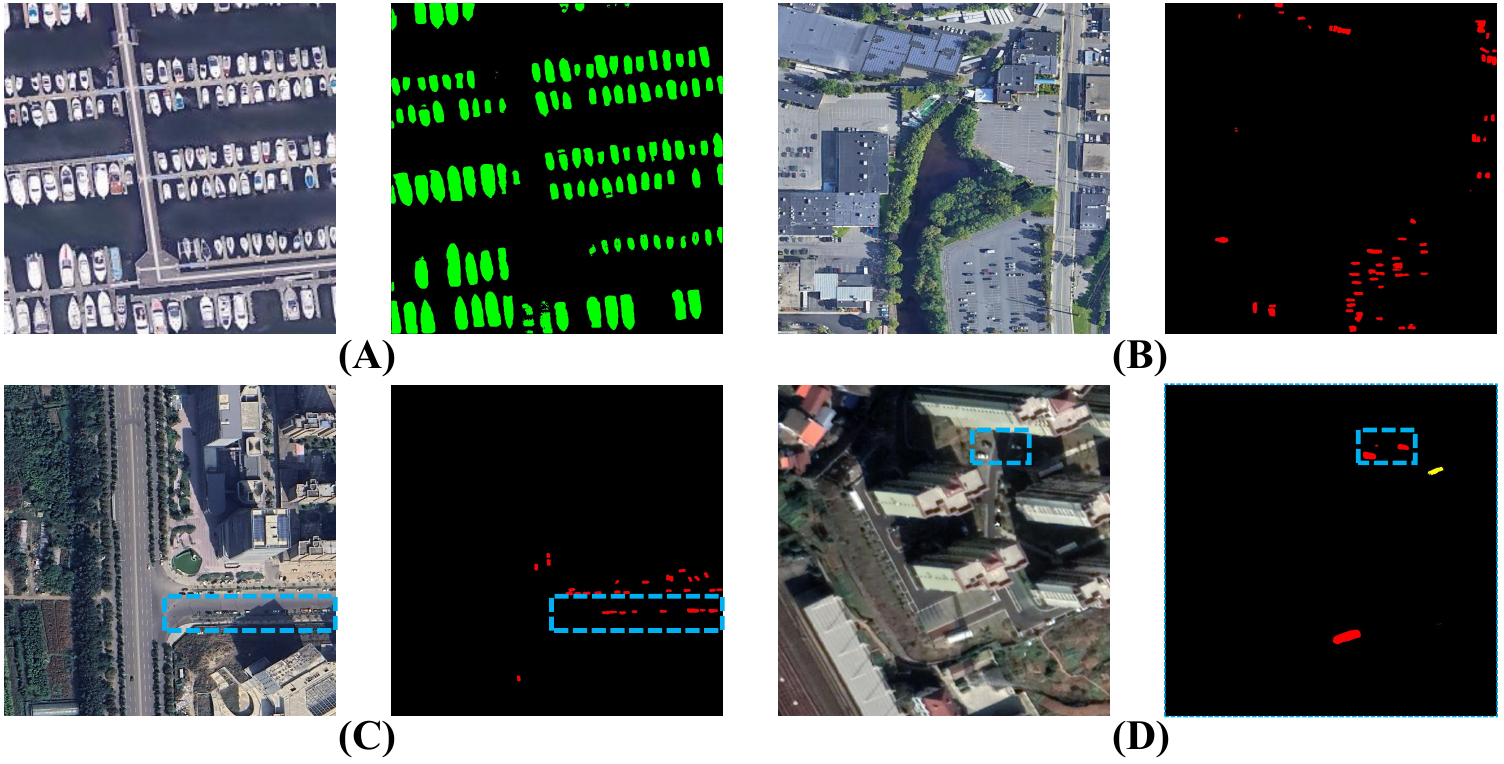}
\caption{Visualization results of the model on the complex scenes.}
\label{fig_10}
\end{figure}
Fig. \ref{fig_10} illustrates the performance of our CSPNet in complex scenes. Subfigures (A) and (B) depict dense environments such as ports and parking. Our model demonstrates the ability to accurately detect and delineate ships and cars within these complex scenes, which produces distinct outlines. This is mainly attributed to the cross spatial-scale feature interaction module, which effectively integrates spatial structural features with deep semantic information. The blue boxes in subfigures (C) and (D) highlight the ability of model to accurately segment cars in shaded or occluded conditions. This is facilitated by the global-local adaptive decoder, which captures and utilizes global context and local features. This mechanism enables the model to perceive overall image characteristics and infer the presence of cars under shadows, thereby maintaining robust performance in challenging lighting conditions.

\begin{table*}[htb]
    \caption{Ablation experiments results on NWPU-Traffic. The best results are in bold, and the second-best results are underlined.}
    \centering
    \footnotesize 
    \setlength{\tabcolsep}{4pt} 
    \renewcommand\arraystretch{1.2} 
    \resizebox{\textwidth}{!}{ 
    \begin{tabular}{lccc|cccccccc}
        \hline
        \hline
        \textbf{Method} & \textbf{SPFIM} & \textbf{CPFIM} & \textbf{LGFFD} & \textbf{Background} $\color{blue}\uparrow$ & \textbf{Airplane} $\color{blue}\uparrow$ & \textbf{Car} $\color{blue}\uparrow$ & \textbf{Ship} $\color{blue}\uparrow$ & \textbf{Train} $\color{blue}\uparrow$ & \textbf{mIoU} $\color{blue}\uparrow$ & \textbf{OA} $\color{blue}\uparrow$ & \textbf{F1-score} $\color{blue}\uparrow$ \\ 
        \hline
        ModelA  & \checkmark &  & \checkmark & \underline{98.62} & \textbf{85.84} & \underline{47.66} & 76.29 & 50.29 & 71.74 & \underline{98.65} & \underline{81.94} \\ 
        ModelB  &  & \checkmark & \checkmark & \underline{98.62} & \textbf{85.84} & 47.47 & \underline{78.53} & 48.55 & \underline{71.80} & \underline{98.65} & 81.88 \\ 

        ModelC  & \checkmark & \checkmark &  & 98.60 & 75.09 & 44.60 & 75.09 & \underline{50.53} & 70.86 & 98.63 & 81.22 \\ 

        CSPNet & \checkmark & \checkmark & \checkmark & \textbf{98.73} & \underline{84.52} & \textbf{48.91} & \textbf{80.07} & \textbf{55.30} & \textbf{73.50} & \textbf{98.76} & \textbf{83.36} \\ 
        \hline
        \hline
        \label{tab3}
    \end{tabular}
    }
\end{table*}
\subsection{Ablation Study}

To thoroughly evaluate the effectiveness of the proposed CPFIM and SPFIM, we conduct comprehensive ablation experiments on the NWPU-Traffic dataset. Specifically, we evaluate four model variants: (1) the full model incorporating CPFIM, SPFIM, and LGFFD, (2) a variant without SPFIM, (3) a variant without CPFIM, and (4) a variant where LGFFD is replaced with a common convolution decoder.
As depicted in Table \ref{tab3}, the complete model demonstrated superior performance across all evaluation metrics. This highlights the crucial role of both SPFIM and CPFIM in enhancing segmentation accuracy. The removal of SPFIM results in a 1.76\% decrease in mIoU and a 1.42\% decrease in F1-score. This degradation is primarily attributed to the critical role of SPFIM in capturing spatial contextual information and refining object boundary localization. In the absence of SPFIM, the model exhibits noticeable performance degradation in handling blurry boundaries and small target segmentation, particularly evident in the segmentation of cars and ships.
Similarly, the elimination of CPFIM leads to a 1.7\% decrease in mIoU and a 1.48\% decrease in F1-score. The results confirm that SPFIM and CPFIM offer complementary advantages. Combining these modules improves the accuracy and robustness of model segmentation.
The ablation study also reveals that the LGFFD component delivered better performance than conventional decoders. Overall, these findings demonstrated that each component of the model played a distinct and valuable role in achieving competitive segmentation performance on the NWPU-Traffic dataset.

\subsection{Complements with ISAID Datasets on Untrained Data}

\begin{figure*}[ht]
\centering
\includegraphics[width=4.8in]{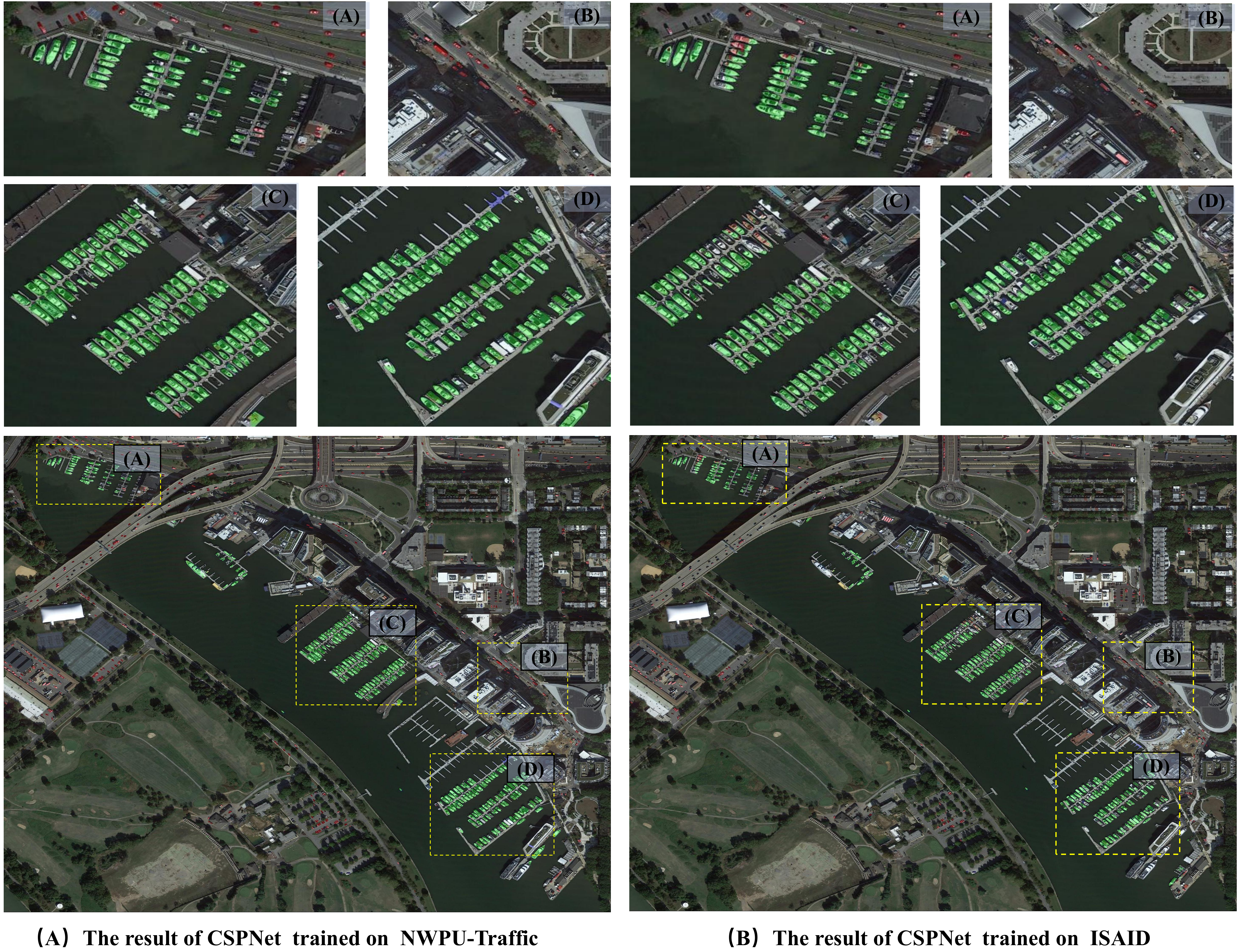}
\caption{Visualization results of the model on the Ports in the District of Columbia, Washington, USA. The left shows inference results from the model trained on the NWPU-Traffic dataset, while the right displays inference results from the model trained on the ISAID dataset.}
\label{fig_11}
\end{figure*}

To evaluate the effectiveness of our proposed NWPU-Traffic dataset, we randomly select a remote sensing image for inference. Importantly, this evaluation dataset is carefully curated to ensure zero overlap with the training distribution, which enables a comprehensive assessment of both the generalization abilities of dataset and its operational effectiveness in realistic scenes. Specifically, we train our CSPNet model on both the ISAID dataset and the NWPU-Traffic dataset and then perform inference on imagery of ports in the District of Columbia to assess the real-world performance of the model. As shown in Fig. \ref{fig_11} (A, C, D), our results demonstrate that the model trained on NWPU-Traffic shows greater accuracy in segmenting docked ships within the dense port area. In contrast, the model trained on ISAID exhibits instances of undetected ships and unclear contours. Additionally, as shown in Fig. \ref{fig_11} (B), in areas where vehicles are shaded by buildings, the NWPU-Traffic-trained model outperforms the other. Overall, these results confirm that our dataset better reflects the characteristics of transportation vehicles and their relevant features in different scenes.

\subsection{Some Visualization Results on Untrained Data
}

To rigorously evaluate the practical deployment performance of our model, we conduct extensive testing on some large-scale real-world remote sensing imagery sourced from diverse geographical regions. 
\begin{figure*}[h]
\centering
\includegraphics[width=4.8in]{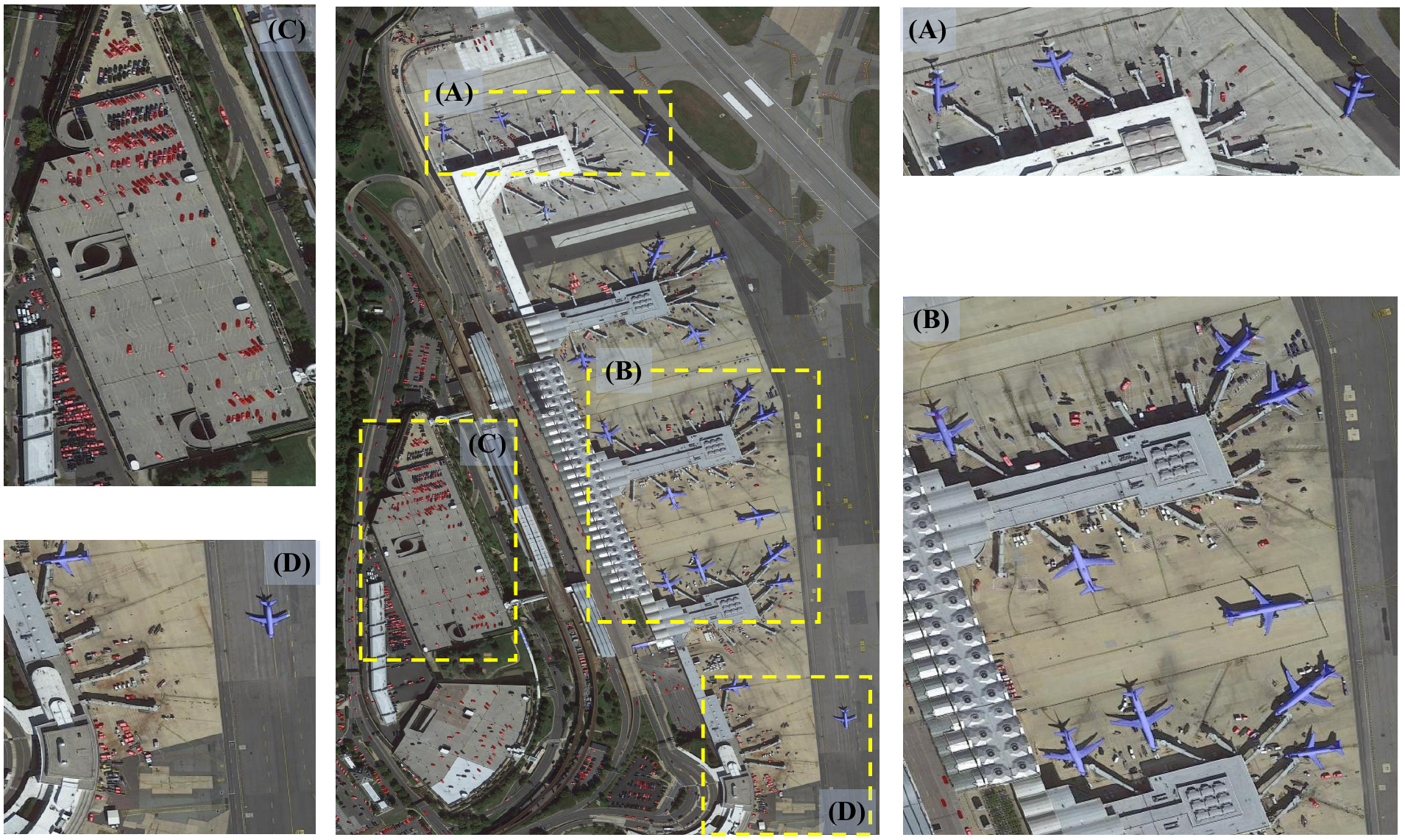}
\caption{Visualization results of the model on the Ronald Reagan Washington National Airport, State of Washington, USA.}
\label{fig_12}
\end{figure*}

Fig. \ref{fig_12} presents remote sensing imagery of Ronald Reagan Washington National Airport, which showcases the ability of our model to accurately segment various transportation targets, including aircraft and ground service vehicles. As shown in Fig. \ref{fig_12} (A) and (D), our model precisely identifies key aircraft components and vehicle contours. This success is attributed to the feature interaction network can effectively capture and integrate distinctive features across different features. Fig. \ref{fig_12} (B) demonstrates robust performance in densely parked aircraft regions, where clear boundaries between individual aircraft are maintained. This indicates that the feature interaction network successfully extracts fine-grained aircraft features to enhance segmentation accuracy. The local-global feature fusion decoder is particularly valuable in handling complex airport layouts. For example, Fig. \ref{fig_12} (C) shows accurate transportation target segmentation near terminal buildings. The model effectively combines local vehicle features with global airport context, which maintains segmentation consistency despite challenging conditions like dense vehicle activity and intricate architectural structures.

\begin{figure*}[h]
\centering
\includegraphics[width=5.1in]{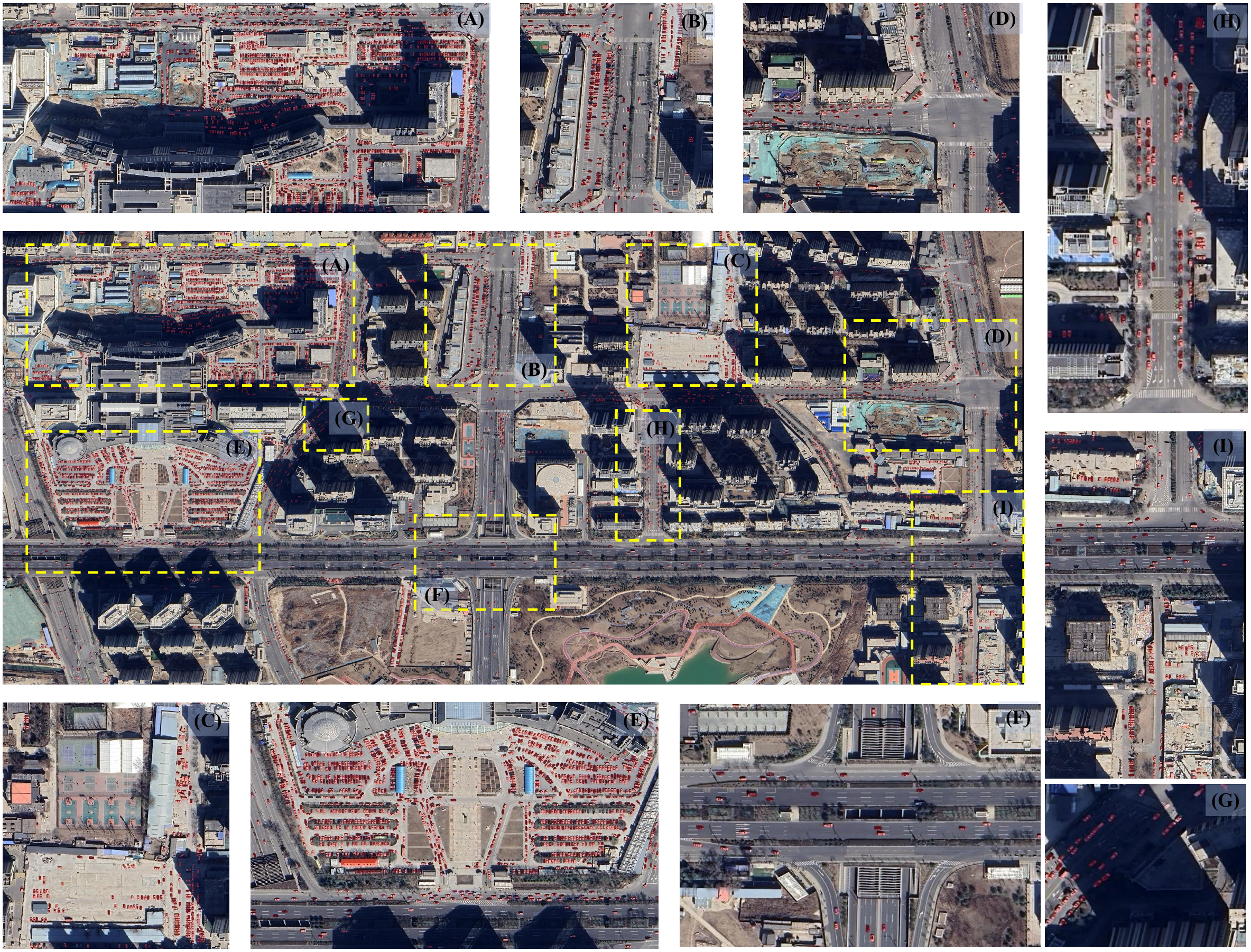}
\caption{Visualization results of the model on the Longcheng street, Taiyuan City, China.}
\label{fig_13}
\end{figure*}

Fig. \ref{fig_13} shows the performance of our CSPNet across diverse urban traffic scenes in Longcheng street, Taiyuan City, China, including arterial roads (A), secondary roads (H), road intersections (B, D, F, I), parking lots (C, E), and building shadow occlusion areas (G). The segmentation results validate the robustness of model in complex real-world conditions. On congested urban roads, the model accurately segments vehicle contours of various types, from compact cars to large buses. The feature interaction network effectively leverages multi-level feature representations. It enables the model to: distinguish subtle differences between adjacent vehicles, maintain segmentation integrity under partial occlusion, and identify individual transportation targets even in densely packed traffic conditions. These abilities significantly enhance the accuracy of urban traffic object segmentation.

\section{Conclusion} \label{sec6}
Precisely identifying and segmenting transportation objects in remote sensing images is both a significant and challenging task. In this paper, we construct a large-scale, multi-scene transportation object segmentation dataset. This dataset encompasses diverse traffic scenes from various countries, which guarantees extensive applicability.  Furthermore, we propose a spatial-channel information preserving segmentation network for transportation objects. The network effectively combines advanced semantic information and positional details from multi-scale features. During the decoding process, it adaptively leverages both local features and global semantic information, significantly enhancing segmentation performance across different scales. Extensive experiments on the proposed dataset demonstrate the superior performance of our method, and ablation studies further validate the efficacy of the design choices. Finally, experiments in several untrained areas show that our dataset and model possess excellent generalization abilities.

In the future, we will work to build a larger traffic dataset, including global ports, airports and typical cities, to more effectively assess traffic conditions in different regions.

\bibliographystyle{elsarticle-num} 
\bibliography{refs.bib}

@article{zhang2022trans4trans,
  title={{Trans4Trans: Efficient transformer for transparent object and semantic scene segmentation in real-world navigation assistance}},
  author={Zhang, Jiaming and Yang, Kailun and Constantinescu, Angela and Peng, Kunyu and M{\"u}ller, Karin and Stiefelhagen, Rainer},
  journal={IEEE Transactions on Intelligent Transportation Systems},
  volume={23},
  number={10},
  pages={19173--19186},
  year={2022},
  publisher={IEEE}
}

@article{liu2016ship,
  title={Ship rotated bounding box space for ship extraction from high-resolution optical satellite images with complex backgrounds},
  author={Liu, Zikun and Wang, Hongzhen and Weng, Lubin and Yang, Yiping},
  journal={IEEE Geoscience and Remote Sensing Letters},
  volume={13},
  number={8},
  pages={1074--1078},
  year={2016},
  publisher={IEEE}
}

@inproceedings{zhu2015orientation,
  title={Orientation robust object detection in aerial images using deep convolutional neural network},
  author={Zhu, Haigang and Chen, Xiaogang and Dai, Weiqun and Fu, Kun and Ye, Qixiang and Jiao, Jianbin},
  booktitle={Proceedings of the IEEE International Conference on Image Processing
},
  pages={3735--3739},
  year={2015},
  organization={IEEE}
}

@inproceedings{waqas2019isaid,
  title={isaid: A large-scale dataset for instance segmentation in aerial images},
  author={Waqas Zamir, Syed and Arora, Aditya and Gupta, Akshita and Khan, Salman and Sun, Guolei and Shahbaz Khan, Fahad and Zhu, Fan and Shao, Ling and Xia, Gui-Song and Bai, Xiang},
  booktitle={Proceedings of the IEEE/CVF Conference on Computer Vision and Pattern Recognition Workshops},
  pages={28--37},
  year={2019}
}

@article{yang2024c,
  title={{C 2 Net: Road Extraction via Context Perception and Cross Spatial-Scale Feature Interaction}},
  author={Yang, Zhigang and Zhang, Wei and Li, Qiang and Ni, Weiping and Wu, Junzheng and Wang, Qi},
  journal={IEEE Transactions on Geoscience and Remote Sensing},
  year={2024},
  publisher={IEEE}
}

@article{cheng2016learning,
  title={{Learning rotation-invariant convolutional neural networks for object detection in VHR optical remote sensing images}},
  author={Cheng, Gong and Zhou, Peicheng and Han, Junwei},
  journal={IEEE Transactions on Geoscience and Remote Sensing},
  volume={54},
  number={12},
  pages={7405--7415},
  year={2016},
  publisher={IEEE}
}

@inproceedings{sun2025frequency,
  title={Frequency-spatial entanglement learning for camouflaged object detection},
  author={Sun, Yanguang and Xu, Chunyan and Yang, Jian and Xuan, Hanyu and Luo, Lei},
  booktitle={Proceedings of European Conference on Computer Vision},
  pages={343--360},
  year={2025},
  organization={Springer}
}

@article{wu2024cmlformer,
  title={{CMLFormer: CNN and Multi-scale Local-context Transformer network for remote sensing images semantic segmentation}},
  author={Wu, Honglin and Zhang, Min and Huang, Peng and Tang, Wenlong},
  journal={IEEE Journal of Selected Topics in Applied Earth Observations and Remote Sensing},
  year={2024},
  publisher={IEEE}
}

@article{huang2022scaleformer,
  title={ScaleFormer: revisiting the transformer-based backbones from a scale-wise perspective for medical image segmentation},
  author={Huang, Huimin and Xie, Shiao and Lin, Lanfen and Iwamoto, Yutaro and Han, Xianhua and Chen, Yen-Wei and Tong, Ruofeng},
  journal={arXiv:2207.14552},
  year={2022}
}

@inproceedings{chen2018encoder,
  title={Encoder-decoder with atrous separable convolution for semantic image segmentation},
  author={Chen, Liang-Chieh and Zhu, Yukun and Papandreou, George and Schroff, Florian and Adam, Hartwig},
  booktitle={Proceedings of European Conference on Computer Vision},
  pages={801--818},
  year={2018}
}

@article{zhu2021global,
  title={A global context-aware and batch-independent network for road extraction from VHR satellite imagery},
  author={Zhu, Qiqi and Zhang, Yanan and Wang, Lizeng and Zhong, Yanfei and Guan, Qingfeng and Lu, Xiaoyan and Zhang, Liangpei and Li, Deren},
  journal={ISPRS Journal of Photogrammetry and Remote Sensing},
  volume={175},
  pages={353--365},
  year={2021},
  publisher={Elsevier}
}

@inproceedings{zhang2024mrfs,
  title={{MRFS: Mutually Reinforcing Image Fusion and Segmentation}},
  author={Zhang, Hao and Zuo, Xuhui and Jiang, Jie and Guo, Chunchao and Ma, Jiayi},
  booktitle={Proceedings of the IEEE/CVF Conference on Computer Vision and Pattern Recognition},
  pages={26974--26983},
  year={2024}
}

@inproceedings{lin2023scale,
  title={Scale-aware modulation meet transformer},
  author={Lin, Weifeng and Wu, Ziheng and Chen, Jiayu and Huang, Jun and Jin, Lianwen},
  booktitle={Proceedings of the IEEE/CVF International Conference on Computer Vision},
  pages={6015--6026},
  year={2023}
}

@inproceedings{zheng2020foreground,
  title={Foreground-aware relation network for geospatial object segmentation in high spatial resolution remote sensing imagery},
  author={Zheng, Zhuo and Zhong, Yanfei and Wang, Junjue and Ma, Ailong},
  booktitle={Proceedings of the IEEE Conference on Computer Vision and Pattern Recognition
},
  pages={4096--4105},
  year={2020}
}

@article{zhu2024review,
  title={A review of multi-class change detection for satellite remote sensing imagery},
  author={Zhu, Qiqi and Guo, Xi and Li, Ziqi and Li, Deren},
  journal={Geo-spatial Information Science},
  volume={27},
  number={1},
  pages={1--15},
  year={2024},
  publisher={Taylor \& Francis}
}

@article{li2024edge,
  title={{Edge-Guided Perceptual Network for Infrared Small Target Detection}},
  author={Li, Qiang and Zhang, Mingwei and Yang, Zhigang and Yuan, Yuan and Wang, Qi},
  journal={IEEE Transactions on Geoscience and Remote Sensing},
  year={2024},
  publisher={IEEE}
}

@inproceedings{guo2024cpp,
  title={{CPP-Net: Embracing Multi-Scale Feature Fusion into Deep Unfolding CP-PPA Network for Compressive Sensing}},
  author={Guo, Zhen and Gan, Hongping},
  booktitle={Proceedings of the IEEE/CVF Conference on Computer Vision and Pattern Recognition},
  pages={25086--25095},
  year={2024}
}

@article{yang2022road,
  title={Road extraction from satellite imagery by road context and full-stage feature},
  author={Yang, Zhigang and Zhou, Daoxiang and Yang, Ying and Zhang, Jiapeng and Chen, Zehua},
  journal={IEEE Geoscience and Remote Sensing Letters},
  volume={20},
  pages={1--5},
  year={2022},
  publisher={IEEE}
}

@article{ren2024pointobb,
  title={Pointobb-v2: Towards simpler, faster, and stronger single point supervised oriented object detection},
  author={Ren, Botao and Yang, Xue and Yu, Yi and Luo, Junwei and Deng, Zhidong},
  journal={arXiv preprint arXiv:2410.08210},
  year={2024}
}

@inproceedings{heitz2008learning,
  title={Learning spatial context: Using stuff to find things},
  author={Heitz, Geremy and Koller, Daphne},
  booktitle={Proceedings of the European Conference on Computer Vision
},
  pages={30--43},
  year={2008},
  organization={Springer}
}

@article{razakarivony2016vehicle,
  title={Vehicle detection in aerial imagery: A small target detection benchmark},
  author={Razakarivony, Sebastien and Jurie, Frederic},
  journal={Journal of Visual Communication and Image Representation},
  volume={34},
  pages={187--203},
  year={2016},
  publisher={Elsevier}
}

@article{lam2018xview,
  title={xview: Objects in context in overhead imagery},
  author={Lam, Darius and Kuzma, Richard and McGee, Kevin and Dooley, Samuel and Laielli, Michael and Klaric, Matthew and Bulatov, Yaroslav and McCord, Brendan},
  journal={arXiv preprint arXiv:1802.07856},
  year={2018}
}

@inproceedings{xia2018dota,
  title={{DOTA}: A large-scale dataset for object detection in aerial images},
  author={Xia, Gui-Song and Bai, Xiang and Ding, Jian and Zhu, Zhen and Belongie, Serge and Luo, Jiebo and Datcu, Mihai and Pelillo, Marcello and Zhang, Liangpei},
  booktitle={Proceedings of the IEEE Conference on Computer Vision and Pattern Recognition
},
  pages={3974--3983},
  year={2018}
}

@article{xiao2015elliptic,
  title={Elliptic Fourier transformation-based histograms of oriented gradients for rotationally invariant object detection in remote-sensing images},
  author={Xiao, Zhifeng and Liu, Qing and Tang, Gefu and Zhai, Xiaofang},
  journal={International Journal of Remote Sensing},
  volume={36},
  number={2},
  pages={618--644},
  year={2015},
  publisher={Taylor \& Francis}
}

@article{zhang2019hierarchical,
  title={Hierarchical and robust convolutional neural network for very high-resolution remote sensing object detection},
  author={Zhang, Yuanlin and Yuan, Yuan and Feng, Yachuang and Lu, Xiaoqiang},
  journal={IEEE Transactions on Geoscience and Remote Sensing},
  volume={57},
  number={8},
  pages={5535--5548},
  year={2019},
  publisher={IEEE}
}

@article{zou2017random,
  title={Random access memories: A new paradigm for target detection in high resolution aerial remote sensing images},
  author={Zou, Zhengxia and Shi, Zhenwei},
  journal={IEEE Transactions on Image Processing},
  volume={27},
  number={3},
  pages={1100--1111},
  year={2017},
  publisher={IEEE}
}

@article{zhang2021sar,
  title={SAR ship detection dataset (SSDD): Official release and comprehensive data analysis},
  author={Zhang, Tianwen and Zhang, Xiaoling and Li, Jianwei and Xu, Xiaowo and Wang, Baoyou and Zhan, Xu and Xu, Yanqin and Ke, Xiao and Zeng, Tianjiao and Su, Hao and others},
  journal={Remote Sensing},
  volume={13},
  number={18},
  pages={3690},
  year={2021},
  publisher={MDPI}
}

@inproceedings{mundhenk2016large,
  title={A large contextual dataset for classification, detection and counting of cars with deep learning},
  author={Mundhenk, T Nathan and Konjevod, Goran and Sakla, Wesam A and Boakye, Kofi},
  booktitle={Proceedings of the European Conference on Computer Vision
},
  pages={785--800},
  year={2016},
  organization={Springer}
}

@article{li2021abcnet,
  title={{ABCNet}: Attentive bilateral contextual network for efficient semantic segmentation of Fine-Resolution remotely sensed imagery},
  author={Li, Rui and Zheng, Shunyi and Zhang, Ce and Duan, Chenxi and Wang, Libo and Atkinson, Peter M},
  journal={ISPRS Journal of Photogrammetry and Remote Sensing},
  volume={181},
  pages={84--98},
  year={2021},
  publisher={Elsevier}
}

@article{wang2022unetformer,
  title={{UNetFormer}: A UNet-like transformer for efficient semantic segmentation of remote sensing urban scene imagery},
  author={Wang, Libo and Li, Rui and Zhang, Ce and Fang, Shenghui and Duan, Chenxi and Meng, Xiaoliang and Atkinson, Peter M},
  journal={ISPRS Journal of Photogrammetry and Remote Sensing},
  volume={190},
  pages={196--214},
  year={2022},
  publisher={Elsevier}
}

@misc{AirbusShipDetectionChallengeDataset,
  title = {Airbus Ship Detection Challenge Dataset},
  howpublished = {\url{https://www.kaggle.com/c/airbus-ship-detection/}},
  year = {2018},
  note = {Accessed: 2019-05-27}
}

@article{yang2024scene,
  title={Scene Adaptive Building Individual Segmentation Based on Large-Scale Airborne LiDAR Point Clouds},
  author={Yang, Wangshan and Zhang, Yongjun and Liu, Xinyi and Gao, Boyong},
  journal={IEEE Transactions on Geoscience and Remote Sensing},
  year={2024},
  publisher={IEEE}
}

@article{wu2023cmtfnet,
  title={{CMTFNet}: CNN and multiscale transformer fusion network for remote-sensing image semantic segmentation},
  author={Wu, Honglin and Huang, Peng and Zhang, Min and Tang, Wenlong and Yu, Xinyu},
  journal={IEEE Transactions on Geoscience and Remote Sensing},
  volume={61},
  pages={1--12},
  year={2023},
  publisher={IEEE}
}

@article{zeng2024multiscale,
  title={Multiscale global context network for semantic segmentation of high-resolution remote sensing images},
  author={Zeng, Qiaolin and Zhou, Jingxiang and Tao, Jinhua and Chen, Liangfu and Niu, Xuerui and Zhang, Yumeng},
  journal={IEEE Transactions on Geoscience and Remote Sensing},
  volume={62},
  pages={1--13},
  year={2024},
  publisher={IEEE}
}

@ARTICLE{10969832,
  author={Li, Langping and Yi, Jizheng and Fan, Hui and Lin, Hui},
  journal={IEEE Transactions on Geoscience and Remote Sensing}, 
  title={A Lightweight Semantic Segmentation Network Based on Self-attention Mechanism and State Space Model for Efficient Urban Scene Segmentation}, 
  year={2025},
  volume={},
  number={},
  pages={},
  doi={10.1109/TGRS.2025.3562185}}

@article{li2023rgb,
  title={{RGB-induced} feature modulation network for hyperspectral image super-resolution},
  author={Li, Qiang and Gong, Maoguo and Yuan, Yuan and Wang, Qi},
  journal={IEEE Transactions on Geoscience and Remote Sensing},
  volume={61},
  pages={1--11},
  year={2023},
  publisher={IEEE}
}

@article{li2023multiscale,
  title={Multiscale factor joint learning for hyperspectral image super-resolution},
  author={Li, Qiang and Yuan, Yuan and Wang, Qi},
  journal={IEEE Transactions on Geoscience and Remote Sensing},
  volume={61},
  pages={1--10},
  year={2023},
  publisher={IEEE}
}

@article{chen2024improving,
  title={Improving image segmentation with contextual and structural similarity},
  author={Chen, Xiaoyang and Liu, Qin and Deng, Hannah H and Kuang, Tianshu and Lin, Henry Hung-Ying and Xiao, Deqiang and Gateno, Jaime and Xia, James J and Yap, Pew-Thian},
  journal={Pattern Recognition},
  volume={152},
  pages={110489},
  year={2024},
  publisher={Elsevier}
}

@article{yang2025dia,
  title={DIA: Deriving linguistic information from auxiliary languages for remote sensing image captioning},
  author={Yang, Tao and Zhou, Qing and Wang, Qi},
  journal={Pattern Recognition},
  pages={112209},
  year={2025},
  publisher={Elsevier}
}

@article{zhan2024yolopx,
  title={YOLOPX: Anchor-free multi-task learning network for panoptic driving perception},
  author={Zhan, Jiao and Luo, Yarong and Guo, Chi and Wu, Yejun and Meng, Jiawei and Liu, Jingnan},
  journal={Pattern Recognition},
  volume={148},
  pages={110152},
  year={2024},
  publisher={Elsevier}
}




\end{document}